\documentclass{article}

\usepackage{arxiv}

\usepackage[utf8]{inputenc} 
\usepackage[T1]{fontenc}    
\usepackage{hyperref}       
\usepackage{url}            
\usepackage{booktabs}       
\usepackage{amsfonts}       
\usepackage{nicefrac}       
\usepackage{microtype}      
\usepackage{lipsum}
\usepackage{amsmath}
\usepackage{array}
\usepackage{graphicx}
\graphicspath{ {./images/} }
\usepackage{subcaption}
 \usepackage[table]{xcolor}
\newcolumntype{P}[1]{>{\centering\arraybackslash}p{#1}}

\title{Benchmarking and Comparing Popular Visual SLAM Algorithms}

\author{
  Amey Kasar \\
  Department of Electronics and Telecommunication Engineering\\
  Pune Institute of Computer Technology\\
  Pune, India \\
  \texttt{kasaramey16@gmail.com} \\
   }

\begin{document}
\maketitle
\begin{abstract}
This paper contains the performance analysis and benchmarking of two popular visual SLAM Algorithms: RGBD-SLAM and RTABMap. The dataset used for the analysis is the TUM RGBD Dataset from the Computer Vision Group at TUM. The dataset selected has a large set of image sequences from a Microsoft Kinect RGB-D sensor with highly accurate and time-synchronized ground truth poses from a motion capture system. The test sequences selected depict a variety of problems and camera motions faced by Simultaneous Localization and Mapping (SLAM) algorithms for the purpose of testing the robustness of the algorithms in different situations. The evaluation metrics used for the comparison are Absolute Trajectory Error (ATE) and Relative Pose Error (RPE). The analysis involves comparing the Root Mean Square Error (RMSE) of the two metrics and the processing time for each algorithm. This paper serves as an important aid in the selection of SLAM algorithm for different scenes and camera motions. The analysis helps to realize the limitations of both SLAM methods. This paper also points out some underlying flaws in the used evaluation metrics.
\end{abstract}


\section{Introduction}
Robotic applications require modeling environment for various tasks, guidance, search and rescue, etc. A precondition for an autonomous robot is to obtain an accurate model of its environment. A major problem in obtaining this model is the problem of pose uncertainty. The  mobile  robot  mapping  problem  under  pose  uncertainty  is  often referred to as the simultaneous localization and mapping (SLAM) or concurrent mapping and localization (CML) problem \cite{smith1986representation,dissanayake2000computationally,montemerlo2003simultaneous}.

SLAM is a complex problem because a robot needs a homogeneous map inorder to localize itself. To obtain a map, the robot requires a good estimate of its location. The limited range and limited field of view of ranging sensor adds to the problems. The mutual dependency among the pose and the map estimates makes the SLAM problem hard. Even a small error in the map could prevent a robot from working in the environment. Hence it is important to tackle the SLAM problem.

Many different techniques to tackle the SLAM problem have been presented. There are different approaches to the problem like Extended Kalman Filter SLAM (EKF), Sparse Extended Information Filter (SEIF), Extended Information Form (EIF), FastSLAM, GraphSLAM. There are also some proposed evaluation metrics \cite{sturm12iros,kummerle2009measuring,steinbrucker2011real,geiger2012we} for comparing the results of different SLAM algorithms. People often use visual inspection to compare maps or overlays with blueprints of buildings for grid-based estimation techniques. This kind of traditional evaluation becomes more and more difficult as new SLAM algorithms show increasing capabilities. Meaningful comparisons between different SLAM algorithms require some common performance metrics. The metrics should enable the user to compare the outcome of different mapping approaches when applying them on the same dataset. This research uses two such common evaluation metrics on the results obtained from popular SLAM algorithms.

This paper utilizes the metrics for evaluation proposed in \cite{sturm12iros} to evaluate the performance of two popular SLAM methods: RGB-D SLAM\cite{endres2012evaluation} and RTABMap\cite{labbe2011memory}. The evaluation metrics used for the comparison are Absolute Trajectory Error (ATE) and Relative Pose Error (RPE).

The TUM RGB-D dataset \cite{TUMRGBD} is applied to both the algorithms and the resulting trajectory estimate is compared to the groundtruth by evaluating Absolute Trajectory Error (ATE), Relative Pose Error (RPE) and the time taken to process the sequence of images. The sequences are intentionally selected to depict the difficulties encountered by SLAM algorithms when operating in real world. The results of the evaluation are analyzed to determine the ideal algorithm for different situations.

\section{System Configuration}

The system used for testing has the following configuration
		\setlength{\tabcolsep}{20pt}
	    \setlength{\arrayrulewidth}{0.2mm}

		\begin{table}[h!]
		  \centering
	    {\begin{tabular}{ |P{5cm}P{5cm}|  }
		
		\hline

	CPU           &    Intel Core i7-7700HQ 2.80GHz x 8    \\ 
	
	Memory           & 8GB     \\ 
	
	Operating System  & Ubuntu 14.04 LTS  \\ 
	
	OS Type          & 64-bit  \\ 
	
	GPU           & GeForce GTX 1050 Ti/PCIe/SSE2       \\
	
	Cuda           & 8.0 \\
    
    SURF         &   enabled                \\
	
	SIFT         & enabled       \\                       
	\hline
	\end{tabular}}
	  \newline\newline
  \caption{System Configuration}\label{tab1}
	\end{table}
\section{Algorithms Selected for Evaluation}
\subsection{RGB-D SLAM}
RGB-D SLAM \cite{endres2012evaluation} works well with a hand-held Kinect-style depth sensor. It uses visual features s.a. SURF or SIFT to match pairs of acquired images, and uses RANSAC to robustly estimate the 3D transformation between them. To achieve online processing, the current image is matched only versus a subset of the previous images. Subsequently, it constructs a graph whose nodes correspond to camera views and whose edges correspond to the estimated 3D transformations. The graph is then optimized with HOG-Man \cite{grisetti2010hierarchical} to reduce the accumulated pose errors.
\subsection{RTABMap}
RTAB-Map (Real-Time Appearance-Based Mapping)\cite{labbe2011memory} is a RGB-D, Stereo and Lidar Graph-Based SLAM approach based on an incremental appearance-based loop closure detector. The loop closure detector uses a bag-of-words approach to determinate how likely a new image comes from a previous location or a new location. When a loop closure hypothesis is accepted, a new constraint is added to the map’s graph, then a graph optimizer minimizes the errors in the map. A memory management approach described in \cite{labbe2011memory} is used to limit the number of locations used for loop closure detection and graph optimization, so that real-time constraints on large-scale environments are always respected. RTAB-Map can be used alone with a handheld Kinect, a stereo camera or a 3D lidar for 6DoF mapping, or on a robot equipped with a laser rangefinder for 3DoF mapping.
\section{Dataset}
 
The TUM RGBD dataset \cite{TUMRGBD} is a large set of data with sequences containing both RGB-D data and ground truth pose estimates from a motion capture system.
The following seven sequences used in this analysis depict different situations and intended to test robustness of algorithms in these conditions.
\subsection{freiburg2 desk with person}
This sequence is a typical office scene with a person sitting at a desk. The person moves and interacts with some of the objects on the table. This sequence is intended for checking the robustness of a SLAM system against dynamic objects and persons, but it can also be used to differentiate maps and find changes in the scene. 
\begin{figure}
\begin{subfigure}{.5\textwidth}
  \centering
  \includegraphics[width=.8\linewidth]{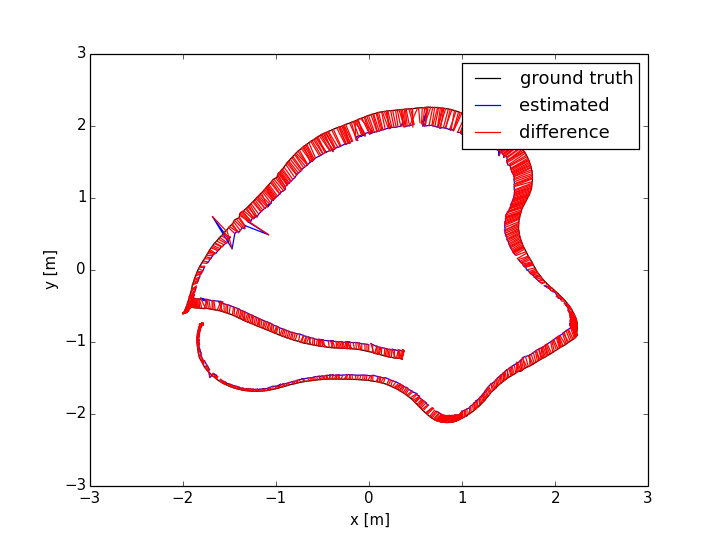}
  \caption{Visualization of Absolute Trajectory Error (ATE) of the estimated trajectory wrt ground truth trajectory}
  \label{fig:sfig1}
\end{subfigure}
\begin{subfigure}{.5\textwidth}
  \centering
  \includegraphics[width=.8\linewidth]{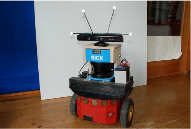}
  \caption{ActiveMedia Pioneer 3 robot}
  \label{fig:sfig2}
\end{subfigure}
\caption{}
\label{fig:fig}
\end{figure}
\subsection{freiburg2 pioneer 360 robot slam}
This sequence was recorded from a Kinect mounted on top of an ActiveMedia Pioneer 3 robot (See Figure \ref{fig:sfig2}). With these sequences, it becomes possible to demonstrate the applicability of SLAM systems to wheeled robots.  Due  to  the  large  dimensions  of  the  hall,  the  Kinect  could not  observe  the  depth  of  the  distant  walls  for  parts  of the  sequence. Several objects like office  containers, boxes,  and  other  feature-poor  objects are placed through the scene.  As  a consequence,  this  sequence  has  depth,  but  is  highly challenging  for  methods  that  rely  on  distinctive  keypoints.

\subsection{freiburg2 360 kidnap}
In this sequence the Kinect sensor is covered several times while it is pointed to a different location ("kidnap") for testing algorithms that can recover from tracking problems. 

\subsection{freiburg3 long office household}
The RGB-D sensor is moved along a large round through a household and office scene with much texture and structure. The end of the trajectory overlaps with the beginning so that there is a large loop closure. 

\subsection{freiburg3 no structure no texture near with loop}
The RGB-D sensor is moved in approximately one meter height along a planar, wooden surface of approximately 3m x 3m. This sequence intentionally has little to no visible structure and texture. The beginning and the end of the sequence overlaps, i.e., there is a loop closure.

\subsection{freiburg3 no structure texture near with loop}
The RGB-D sensor has been moved in one meter height in a circle a textured, planar surface. The texture is highly discriminative as it consists of several conference posters. The beginning and the end of the trajectory overlap, so that there is a loop closure. 

\subsection{freiburg3 sitting static}
In this sequence, two persons sit at a desk, talk, and gesticulate. The sensor has been kept in place manually. This sequence is intended to evaluate the robustness of visual SLAM and odometry algorithms to slowly moving dynamic objects. 
\bigbreak
\section{Evaluation Metrics}
 
The output of a SLAM algorithm is an estimated camera trajectory along with an estimate of the resulting map. While it is in principle possible to evaluate the quality of the resulting map, accurate ground truth maps are impossible to obtain as there are various uncontrollable factors involved. So we base our analysis on the quality of the estimated trajectory obtained from a sequence of RGB images. Even though an accurate trajectory does not imply a good map and error free operation of robot, it is the most common parameter used to measure the accuracy of SLAM algorithms. For the evaluation, we assume that the output of the algorithm is a sequence of poses from the estimated trajectory $P_1,....,P_n \in SE$ and from the ground truth trajectory $Q_1,....,Q_n \in SE$. \\

For simplicity of notation, it is assumed that the two sequences are time-synchronized, equally sampled, and both have length $n$. However, in reality, these two sequences vary in sampling rates, lengths and have missing data, so an additional data association and interpolation step is required. Both sequences consist of homogeneous transformation matrices that give the pose of the RGB optical frame of the RGBD sensor from some other arbitrary reference frame. This reference frame does not have to be the same for both sequences, i.e., the estimated sequence might start at the origin, while the ground truth sequence is an absolute coordinate frame which was defined during calibration. While, in principle, the choice of the reference frame on the RGBD sensor is also arbitrary, the RGB optical frame is used as the reference because the depth images in the dataset have already been registered to this frame. In the remainder of this section, two common evaluation metrics for visual odometry and visual SLAM evaluation given in \cite{sturm12iros} are defined.

\subsection{Absolute Trajectory Error (ATE)}
For a visual SLAM system, the global consistency of the estimated trajectory is an important quantity. The Absolute Trajectory Error (ATE) is evaluated by comparing the absolute distances between the estimated and the ground truth trajectory. As both trajectories can be specified in arbitrary coordinate frames, they first need to be aligned. This is done in closed form by using the method of Horn \cite{horn1987closed}, which finds the rigid-body transformation $S$ corresponding to the least-squares solution that maps the estimated trajectory $P_{1:n}$ onto the ground truth trajectory $Q_{1:n}$. The absolute trajectory error at time step $i$ can be calculated as
\begin{equation}
F_i:=Q_i^{-1}SP_i  
\end{equation}
Evaluating the root mean squared error (RMSE) over all time indices of the translation components, we get,
\begin{equation}
  RMSE(F_{1:n}):=\bigg(\frac{1}{n} \sum_{i=1}^{n} \lvert \lvert trans(F_i) \rvert \rvert ^2 \bigg)^{ 1/2}
\end{equation}
\\

A visualization of the absolute trajectory error is given in Figure \ref{fig:sfig2}. Here, RGB-D SLAM \cite{endres2012evaluation} was used to estimate the camera trajectory from the “fr2 360 pioneer slam” sequence.

\subsection{Relative Pose Error (RPE)}
The Relative Pose Error measures the local accuracy of the trajectory over a fixed time interval $ \Delta $. Therefore, the Relative Pose Error corresponds to the drift of the robot's trajectory which is useful for the evaluation of visual odometry systems. We define the relative pose error at time step $i$ as
\begin{equation}
E_{i}:=(Q_i^{-1} Q_{i + \Delta})^{-1}(P_i^{-1} P_{i + \Delta})^{-1}  
\end{equation}
From a sequence of $n$ camera poses, we obtain in this way $ m = n - \Delta $ individual relative pose errors along the sequence. From these errors, we propose to compute the Root Mean Squared Error (RMSE) over all time indices of the translation component as
\begin{equation}
RMSE(E_{1:n},\Delta):=\bigg(\frac{1}{m} \sum_{i=1}^{m} \lvert \lvert trans(E_i) \rvert \rvert ^2 \bigg)^{ 1/2 }  
\end{equation}
where $trans(E_i)$ refers to the translation components of the Relative Pose Error $E_i$. Some prefer to evaluate the mean error instead of the root mean squared error as it affords less influence to outliers. Some use the median instead of the mean, which attributes even less influence to outliers. For visual odometry systems that match consecutive frames,, the time parameter $\Delta$ is $\Delta=1$ which is an intuitive choice; $RMSE (E_{1:n})$ then gives the drift per frame. For systems that use more than one previous frame, larger values of $\Delta$ can also be appropriate, for example, $\Delta = 30$ gives the drift per second on a sequence recorded at 30 Hz. A common choice is to set $\Delta = n$ which means that the start point is directly compared to the end point. This metric can be misleading as it penalizes rotational errors in the beginning of a trajectory more than towards the end \cite{kummerle2009measuring,kelly2004linearized} and must not be used. It therefore makes sense to average over all possible time intervals $\Delta$, i.e., to calculate
\begin{equation}
RMSE(E_{1:n}):=\frac{1}{n} \sum_{\Delta=1}^n RMSE(E_{1:n},\Delta)  
\end{equation}
Note that the computational complexity of this expression is quadratic in the trajectory length. Therefore, it is approximated by calculating it from a fixed number of relative pose samples.
\\

The RPE can be used to evaluate the global error of a trajectory by averaging over all possible time intervals. The RPE assesses both translational and rotational errors, while the ATE only assesses the translational errors. Therefore, the RPE is always greater than the ATE (or equal if there is no rotational error). Thus, the RPE metric gives us a way to combine rotational and translational errors into a single measure. However,
rotational errors are also indirectly captured by the ATE as it manifest itself in wrong translations. From a practical perspective, the ATE has an intuitive visualization which facilitates visual inspection. Nevertheless, the two metrics are strongly correlated.

\section{Results}
 \begin{figure}
\begin{subfigure}{.5\textwidth}
  \centering
  \includegraphics[width=.8\linewidth]{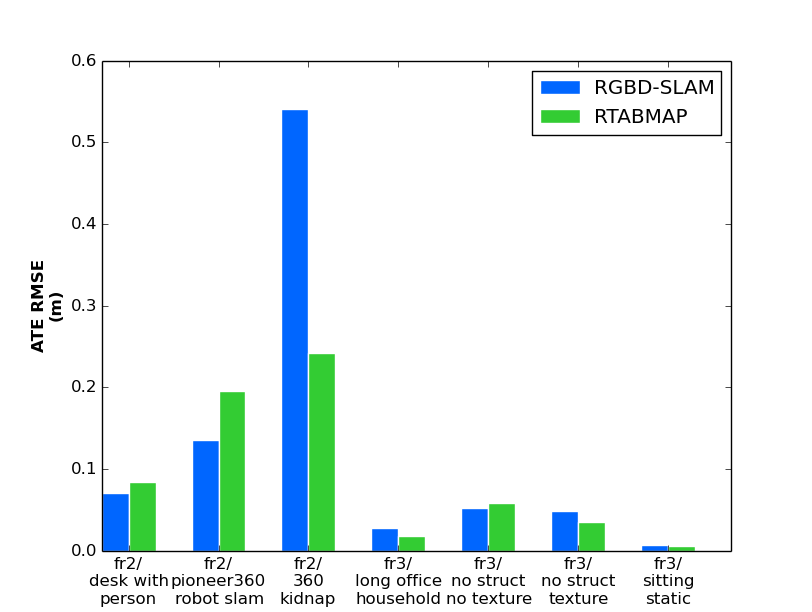}
  \caption{Absolute Trajectory Error (RMSE)}
\end{subfigure}
\begin{subfigure}{.5\textwidth}
  \centering
  \includegraphics[width=.8\linewidth]{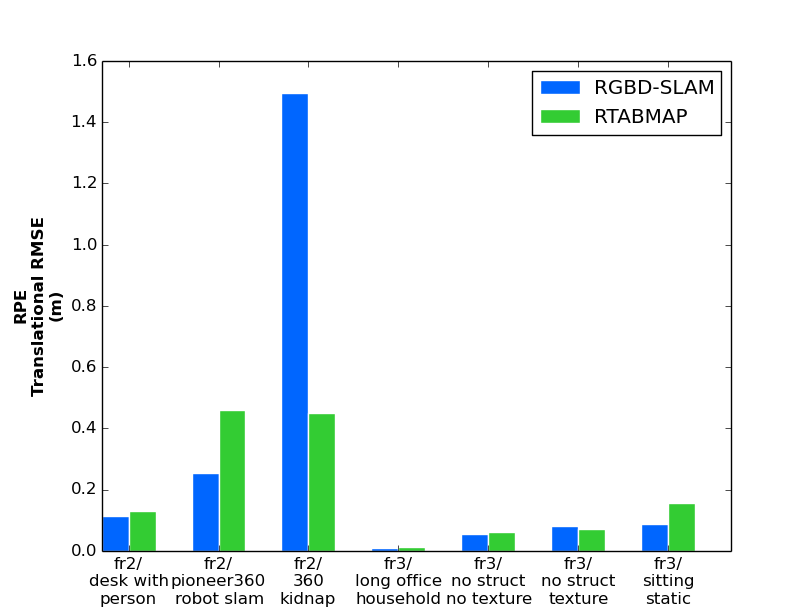}
  \caption{Relative Pose Error Translational (RMSE)}
\end{subfigure}
\begin{subfigure}{.5\textwidth}
  \centering
  \includegraphics[width=.8\linewidth]{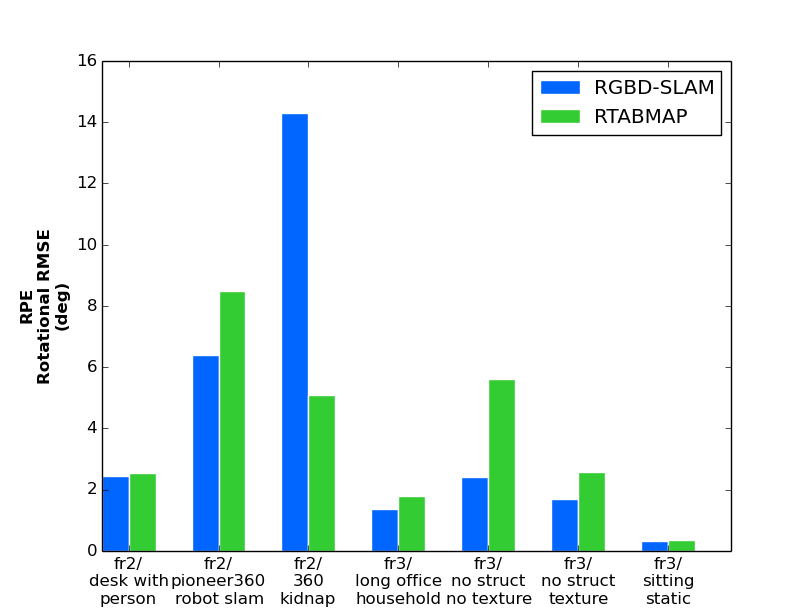}
  \caption{Relative Pose Error Rotational (RMSE)}
\end{subfigure}
\begin{subfigure}{.5\textwidth}
  \centering
  \includegraphics[width=.8\linewidth]{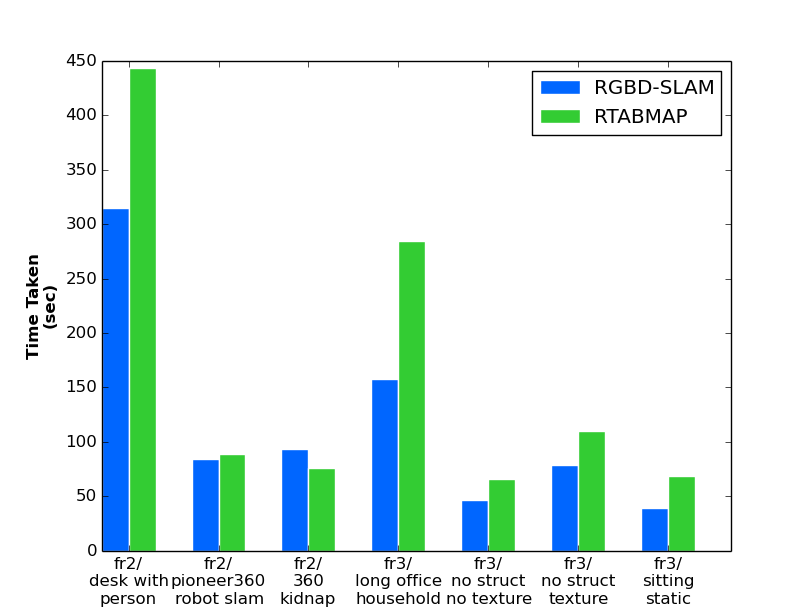}
  \caption{Time taken to process the sequence}
\end{subfigure}
\caption{Result of the benchmark evaluation over seven different sequences}
\label{fig:result}
\end{figure}
The evaluation of both the SLAM methods was carried out on the selected sequences from the dataset. The values of Absolute Trajectory Error, Relative Pose Error are root mean squared. The time required to process the entire sequence was recorded. All the values were plotted for ease of comparison (See Figure \ref{fig:result}). The estimated trajectories and detailed results including min, max, mean, mode, and median values of errors are available at \cite{git}.

\section{Conclusions}
 
\begin{figure}[h!]
\begin{subfigure}{.5\textwidth}
  \centering
  \includegraphics[width=.8\linewidth]{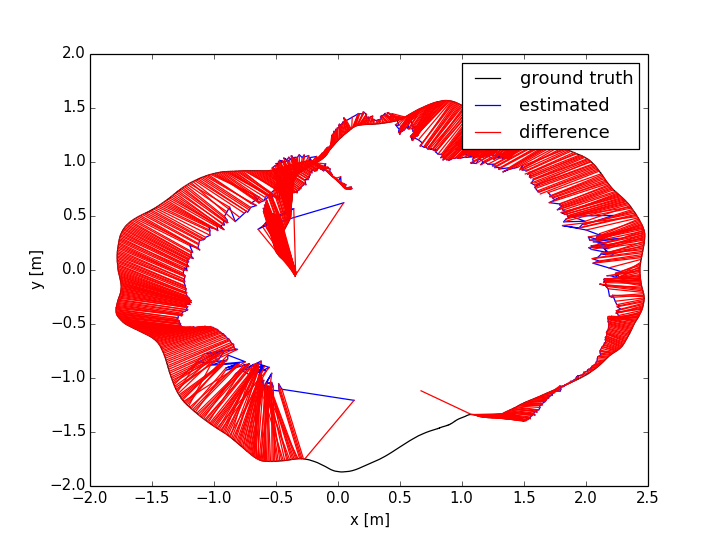}
  \caption{RGB-D SLAM recovers and generates estimated trajectory for the entire motion except when it sensor was covered}
\end{subfigure}
\begin{subfigure}{.5\textwidth}
  \centering
  \includegraphics[width=.8\linewidth]{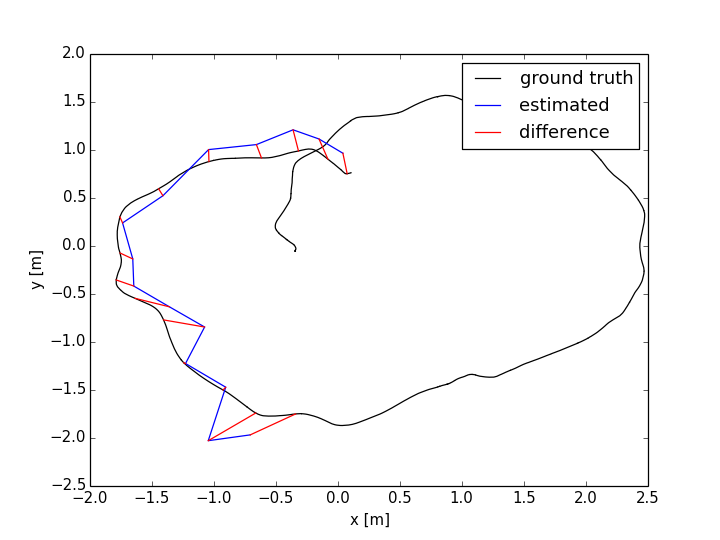}
  \caption{RTABMap fails to generate estimaated trajectory after the sensor was covered}
\end{subfigure}
\caption{ATE visualization of fr2 360 kidnap sequence with RGB-D SLAM and RTABMap}
\label{fig:360kidnap}
\end{figure}
\begin{figure}
\begin{subfigure}{.5\textwidth}
  \centering
  \includegraphics[width=.8\linewidth]{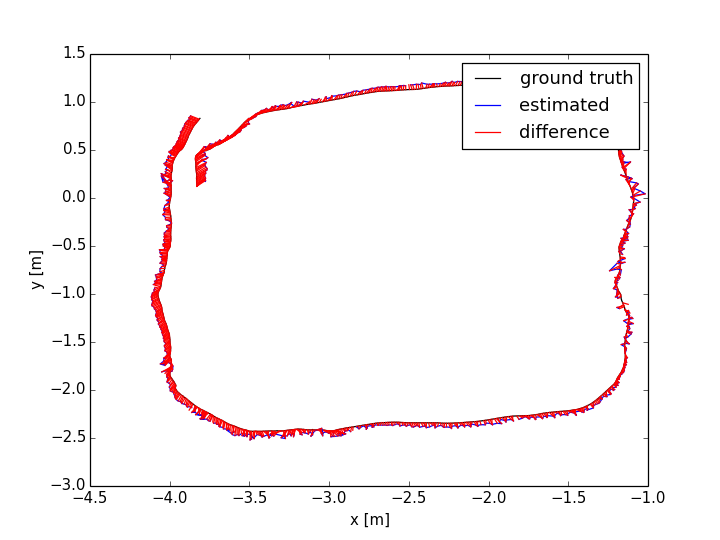}
  \caption{RGB-D SLAM generates estimated trajectory for the entire motion}
\end{subfigure}
\begin{subfigure}{.5\textwidth}
  \centering
  \includegraphics[width=.8\linewidth]{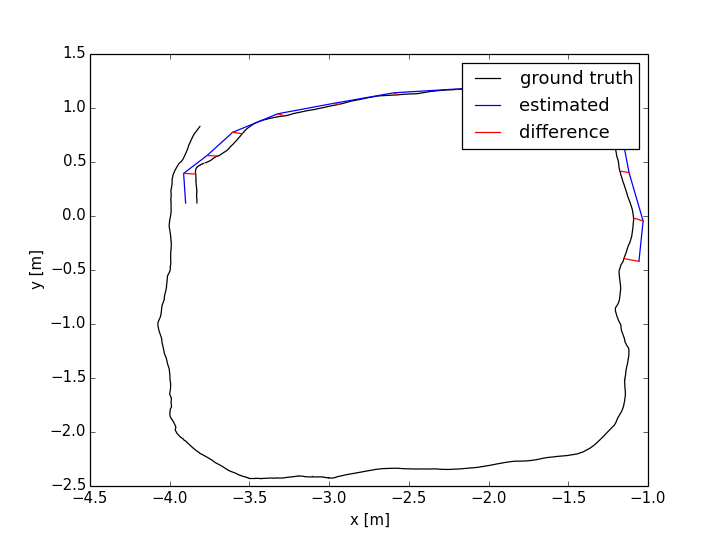}
  \caption{RTABMap fails to generate estimaated trajectory after losing track due to lack of features}
\end{subfigure}
\caption{ATE visualization of fr3 no structure no texture with RGB-D SLAM and RTABMap}
\label{fig:nsnt}
\end{figure}
From the obtained results it can be concluded that RTABMap takes longer to process a sequence than RGB-D SLAM. The exception being fr2 360 kidnap; the reasons for which are discussed later on in this section. A general trend can not be observed from the Absolute Trajectory Error (ATE) RMSE comparison to clearly suggest which SLAM algorithm performs better. We must observe the estimated trajectory from both SLAM algorithms to analyze the problems faced by each.

The estimated trajectory comparison for fr2 360 kidnap sequence when juxtaposed with the groundtruth (See Figure \ref{fig:360kidnap}) reveals that RTABMap failed to recover after the sensor was covered. No estimated trajectory exists for the motion after the covering of the sensor for RTABMap. This is clearly a flaw with the used evaluation metrics as the evaluation ignores the motion for which no poses were generated. RGB-D SLAM recovers from the covering of the sensor and generates poses for the motion afterwards. Because of how the evaluation metrics are defined, RGB-D SLAM could be misidentified as the worse of the two algorithms in this situation despite its recovery from the tracking error.

The time taken by RTABMap to process the sequence is less than the time taken by RGB-D SLAM due to the same reason. RTABMap fails to recover from the tracking problem and does not process the remaining sequence thus resulting in a shorter processing time.

In case of fr3 no structure no texture sequence, the same flaw is observed (See Figure \ref{fig:nsnt}). RTABMap fails to generate valid poses for the latter half of the motion. The algorithm lost tracking due to lack of features in the frame. The algorithm never recovered and hence we have an incomplete estimation of the trajectory.

It must be emphasized that RTABMap generates significantly lesser number of pose pairs for evaluation than RGB-D SLAM. Hence an accurate comparison is not possible with the raw estimated trajectories generated by both methods.

After the analysis of the results it can be concluded that RGB-D SLAM has a better performance overall in the tested circumstances while ignoring the underlying flaws in the evaluation metrics. It has a shorter processing time. It is better at recovering from tracking errors than RTABMap. It also sustains fast camera motion much better than RTABMap.

A need for a better evaluation metric was realized due to the aforementioned flaws in the evaluation metrics as observed from the results.
\section{Acknowledgement}

The author would like to thank Jurgen Sturm from the Computer Vision Group at TUM for without his dataset and benchmark evaluation method this paper would not have been possible. The author also appreciates the support and resources provided by Research Center at Pune Institute of Computer Technology (PICT), India . The author also thanks Dr. Gaurav Bansod and Dr. Geetanjali Kale for their guidance and support.

\end{document}